\documentclass[12pt]{cibb}

\makeatletter
\providecommand{\@ordinalM}[2]{#1}
\makeatother

\usepackage{subfigure,graphicx}
\usepackage{amsmath,amsfonts,latexsym,amssymb,euscript,xr,bm}
\usepackage{booktabs}
\usepackage[nodayofweek]{datetime}
\usepackage{hyperref}
\usepackage{fmtcount}
\usepackage[english]{datenumber}
\usepackage[absolute]{textpos}
\usepackage[table]{xcolor}
\usepackage{color,colortbl,tabularx}
\usepackage[english]{babel}
\usepackage[protrusion=true,expansion=true]{microtype}
\usepackage{amsmath,amsfonts,amsthm}
\usepackage{pifont}

\definecolor{LightBlue}{rgb}{0.88,0.9,0.9}

\title{\Large $\ $\\ \bf Vision4PPG: Emergent PPG Analysis Capability of Vision Foundation Models for Vital Signs like Blood Pressure}

\author{\large Saurabh Kataria, Ayca Ermis, Lovely Yeswanth Panchumarthi, Minxiao Wang, Xiao Hu}
\address{Nell Hodgson Woodruff School of Nursing, Emory University, Atlanta, USA, 30322}

\abstract{\small Photoplethysmography, Foundation Models, Vision Transformers, Non-invasive Blood Pressure, Fine-tuning, Lab Estimation. \normalsize
\\[17pt]
{\bf Abstract.}
Photoplethysmography (PPG) sensor in wearable and clinical devices provides valuable physiological insights in a non-invasive and real-time fashion.
Specialized Foundation Models (FM) or repurposed time-series FMs are used to benchmark physiological tasks.
Our experiments with fine-tuning FMs reveal that Vision FM (VFM) can also be utilized for this purpose and, in fact, surprisingly leads to state-of-the-art (SOTA) performance on many tasks, notably blood pressure estimation.
We leverage VFMs by simply transforming one-dimensional PPG signals into image-like two-dimensional representations, such as the Short-Time Fourier transform (STFT).
Using the latest VFMs, such as DINOv3 and SIGLIP-2, we achieve promising performance on other vital signs and blood lab measurement tasks as well.
Our proposal, Vision4PPG, unlocks a new class of FMs to achieve SOTA performance with notable generalization to other 2D input representations, including STFT phase and recurrence plots.
Our work improves upon prior investigations of vision models for PPG by conducting a comprehensive study, comparing them to state-of-the-art time-series FMs, and demonstrating the general PPG processing ability by reporting results on six additional tasks.
Thus, we provide clinician-scientists with a new set of powerful tools that is also computationally efficient, thanks to Parameter-Efficient Fine-Tuning (PEFT) techniques.
}

\begin{document}


\thispagestyle{myheadings}
\pagestyle{myheadings}

\section{Introduction}
\label{sec:introduction}

Photoplethysmography (PPG) sensors are now ubiquitous, thanks to the widespread adoption of wearable devices~\cite {charlton20232023}.
They are equally crucial in clinical settings, whether in on- or off-hospital data capture scenarios~\cite{angelucci2025wearable}.
Powerful feature extraction of PPG can unlock its true potential, as current error levels (under motion, skin tone, perfusion, and contact pressure variability) remain a significant limitation across many downstream tasks~\cite{elgendi2024recommendations,scardulla2023photoplethysmograhic,charlton20232023}.
Recently, Foundation Models (FMs) trained on physiological modalities or general time-series have been used in conjunction with Parameter-Efficient Fine-Tuning (PEFT) techniques to tune them for PPG tasks~\cite{bommasani2021opportunities,hu2022lora}.
Increasingly, more medical waveform data is being released in the public domain, which necessitates the development of more tools for analysis.
We propose to leverage an existing and overlooked class of FMs: Vision FMs (VFM), by simply transforming PPG to 2D input representations for comprehensive PPG processing.
Prior works on time-series have utilized vision models with features such as line plots~\cite{zeng2023pixels}, spectral features~\cite{dixit2024vision}, and Continuous Wavelet Transform (CWT)~\cite{darbhasayanam2024photoplethysmogram}, Recurrence plots~\cite{hatami2018classification}, Gramian Angular Summation/Difference Fields (GASF/GADF)~\cite{wang2015imaging}, and Markov Transition Fields (MTF)~\cite{wang2015imaging}.

We are interested in physiological signal processing.
In this context, VFMs have been explored to some extent.
In \cite{freitas2023assessing}, 2D representations have yielded much superior signal quality assessment.
\cite{pankaj2023optimized} comes close to our work with spectral features, but does not demonstrate the full potential of the approach to achieve state-of-the-art (SOTA) results.
Limitations of the study include the pursuit of classification of BP rather than regression to absolute values, the use of multi-modal signals, and limited testing with older models.
Our work, in contrast, reports absolute errors on multiple non-invasive BP datasets using just PPGs processed with the latest FMs.
\cite{darbhasayanam2024photoplethysmogram} reports absolute errors using VFMs, but has limited comparison to baselines and uses a single test set.
Our study presents results on multiple datasets and demonstrates a more pronounced improvement over baselines.
Moreover, we selected multiple tasks to demonstrate the \textit{emergent capability of VFMs for generalized PPG signal analysis}.
\cite{li2025continuous} also shows promise of the proposed approach, but lacks sufficient comparisons and suffers from limitations of prior works, like a single-dataset study and non-use of VFMs.
Other domains investigating this include Remote PPG (rPPG)~\cite{choi2024dino} and Electrocardiogram (ECG)~\cite{vaid2023foundational}.

Thus, our unique contributions are:
1) Comprehensive demonstration of pre-trained Vision Foundation Models (unimodal \& multi-modal) for PPG processing and comparison with state-of-the-art time-series foundation models;
2) Curation of multiple PPG test sets with notable improvements in non-invasive blood pressure estimation;
3) Reporting results on more vital tasks like heart rate, respiration rate, SPO2, and blood biomarkers to demonstrate the comprehensive PPG processing and emergent ability of Vision Foundation Models.

\section{Methods and data}
\label{sec:METHODS}

For baseline time-series models, we choose one general time-series model and one PPG domain-specific model.
First, MOMENT~\cite{goswami2024moment}, which is trained on various time-series domains, including weather, finance, and ECG.
Second, PPG-GPT~\cite{chen2025gpt,chen2024adapting}, which is trained on millions of hours of Intensive Care Units (ICU) PPG signals.
We compare these two Time-Series FM (TSFM) with two Vision FMs: DINOv3~\cite{simeoni2025dinov3,liu2025does} and SIGLIP-2~\cite{tschannen2025siglip}.
Vision FMs are an interesting class of FMs that have been demonstrated to estimate multi-channel ECG from PPG~\cite{li5360425beyond} -- suggesting their suitability for converting PPG to patches for Vision Transformer (ViT) processing.
Other candidate models are DINOv2~\cite{oquab2023dinov2} and PEcore~\cite{bolya2025perception}.
We defer their investigation to future work.

DINOv3 is the third generation of the DINO-series VFM by Meta.
It is pretrained (self-supervised) on the curated LVD-1689M web dataset (~1.689B Instagram images).
We use the mid-size variant of ViT-B, with 86M parameters, and an expected patch size of 16x16.
It is derived from a \textit{frozen ViT-7B teacher} with training that includes self-distillation loss, iBOT masked-image modeling, KoLeo regularization (on the CLS), and the new Gram anchoring trick to maintain stable dense features during long training.
DINOv3 has been used for tuning to the medical image domain~\cite{li2025meddinov3,wang2025dinov3}, but not for PPG signals, to the best of our knowledge.

SIGLIP 2 is the second generation of the multi-lingual multi-modal FM by Google.
It combines captioning-based pretraining, self-supervised losses (including self-distillation and masked prediction), and online data curation.
We again use the base 86M variant, similar to DINOv3.
Models are pre-trained on WebLI (10B images paired with 12B alternative texts across 109 languages and more)~\cite{chen2022pali}, and we adopt the fixed-resolution vision tower/encoder as a high-quality, frozen backbone.
SIGLIP has also been used for medical image fine-tuning on cleft lip~\cite{nantha2025enhanced} and histopathology~\cite{gilal2025pathvlm} tasks.
Using the Parameter-Efficient Fine-Tuning (PEFT) technique called Low-Rank Adaptation (LoRA)~\cite{hu2022lora}, we tune the VFMs with hyperparameters $r=8$, $\alpha=16$, and dropout $p_{\text{drop}}=0.05$.
We apply LoRA to the Q, K, and V matrices within the self-attention mechanism.

Focusing on the cuffless/non-invasive Blood Pressure (BP) problem, we curated seven datasets for diastolic and systolic pressure estimation separately.
Our test sets are: 1) PPG-BP~\cite{liang2018new}, 2) Aurora-Oscillometric~\cite{mieloszyk2022comparison}, 3) Aurora-Auscultatory~\cite{mieloszyk2022comparison}, 4) CAS-BP~\cite{liu2023cuffless}, 5) Vital Videos~\cite{toye2023vital}, 6) BCG~\cite{carlson2020bed}, and 7) BUT-PPG~\cite{nemcova2021brno}.
We obtain 30-second 40Hz signals from these datasets, which are derived from a diverse population.
The participants range from 40 to over 2000, while the data size ranges from  0.4 hours to over 170 hours.
The other choices of PPG vitals estimation tasks include heart rate estimation, respiration rate estimation, and SPO2 (blood oxygen saturation).
The datasets we use for these are DALIA~\cite{reiss2019deep}, BIDMC~\cite{pimentel2016toward}, and MIMIC-III~\cite{moody2020mimiciiiwdb} respectively.
Among laboratory estimation tasks, we chose to predict sodium, potassium, and lactate levels using MIMIC-III datasets.
These choices of blood electrolytes cover orthogonal physiological analysis axes and may capture cues for monitoring critical events like sepsis, cardiac arrest, and seizure~\cite{lee2021prognostic,kyaw2022hypokalemia,seethapathy2023severe}.

\section{Input Representations}
Here, we describe the three transforms we use to feed PPGs to the VFM: STFT, STFT+phase, and recurrence plot.
We conduct our preliminary study with these features as they are known to encode relevant information.
STFT can encode bands corresponding to breathing rates~\cite{iqbal2022photoplethysmography} and capture the effects of arterial stiffness~\cite{charlton2022assessing}.
Fourier phase can provide additional timing information in the signals, potentially helping BP estimation tasks~\cite{natarajan2021photoplethysmography}.
We also experiment with recurrence plots (RP)~\cite{dimitriev2020recurrence}, which work on original, derivative, and second derivative~\cite{inoue2017second}, potentially capturing stiffness-related morphology changes with the velocity and acceleration analysis.

\paragraph*{1) STFT (log-power), replicated to 3 channels~\cite{najarian2012biomedical}}
Let $x[n]$ be the discrete-time PPG signal.
The Short-Time Fourier transform (STFT) at frequency bin $f$ and \textit{frame} $t$ is $X(f,t)=\sum_{m=0}^{N_w-1} x[tH+m]\,w[m]\,e^{-j2\pi fm/N_{\mathrm{FFT}}}$, where $w[m]$ is the analysis window of length $N_w$, $H$ is the hop size, and $N_{\mathrm{FFT}}$ is the DFT size; $f\in\{0,\dots,N_{\mathrm{FFT}}-1\}$ and $t$ is the frame index.
$\Re\{\cdot\}$ and $\Im\{\cdot\}$ denote real and imaginary parts.
The phase is $\phi(f,t)=\mathrm{atan2}(\Im X(f,t),\,\Re X(f,t))\in(-\pi,\pi]$.
$\varepsilon>0$ is a small constant used for numerical stability.
The log-power spectrogram is $L(f,t)=\log(|X(f,t)|^2+\varepsilon)$.
We feed a 3-channel image $\{\mathcal{I}_1,\mathcal{I}_2,\mathcal{I}_3\}$ to the vision backbone.
Here, $z$ is the standard z-score normalization.

\begin{equation}
I(f,t)=\mathrm{z}\!\big(L(f,t)\big).
\end{equation}
\begin{equation}
\mathcal{I}_1(f,t)=I(f,t),\quad \mathcal{I}_2(f,t)=I(f,t),\quad \mathcal{I}_3(f,t)=I(f,t).
\end{equation}

\paragraph*{2) STFT + phase (mag + cos phase + sin phase)~\cite{najarian2012biomedical}}
Using previously defined terminology, we use additional phase information here to guide the VFM fine-tuning.
This representation ensures we provide complete information about the signal that the log-power alone misses.
\begin{equation}
\phi(f,t)=\mathrm{atan2}\!\big(\Im X(f,t),\,\Re X(f,t)\big).
\end{equation}
\begin{equation}
\mathcal{I}_1(f,t)=\mathrm{z}\!\left(\log\!\big(|X(f,t)|^2+\varepsilon\big)\right).
\end{equation}
\begin{equation}
\mathcal{I}_2(f,t)=\mathrm{z}\!\big(\cos\phi(f,t)\big).
\end{equation}
\begin{equation}
\mathcal{I}_3(f,t)=\mathrm{z}\!\big(\sin\phi(f,t)\big).
\end{equation}

\paragraph*{3) Recurrence plot~\cite{eckmann1995recurrence}}
Let $\tilde{x}[n]$ be the normalized signal; $d^{(1)}[n]$ is the slope and $d^{(2)}[n]$ is the curvature, which are the first and second discrete differences.
We use the Gaussian-soft recurrence matrix~\cite{marwan2023trends} for a sequence $v$ with bandwidth $\sigma>0$ and time indices $i,j$; the three channels are $\mathcal{I}_k(i,j)$.

\begin{equation}
\tilde{x}[n]=\mathrm{z}\!\big(x[n]\big).
\end{equation}
\begin{equation}
d^{(1)}[n]=\tilde{x}[n]-\tilde{x}[n-1].
\end{equation}
\begin{equation}
d^{(2)}[n]=d^{(1)}[n]-d^{(1)}[n-1].
\end{equation}
\begin{equation}
R_v(i,j)=\exp\!\left(-\frac{(v[i]-v[j])^2}{2\sigma^2}\right),
\quad v\in\{\tilde{x},\,d^{(1)},\,d^{(2)}\}.
\end{equation}
\begin{equation}
\mathcal{I}_1(i,j)=\mathrm{z}\!\big(R_{\tilde{x}}(i,j)\big).
\end{equation}
\begin{equation}
\mathcal{I}_2(i,j)=\mathrm{z}\!\big(R_{d^{(1)}}(i,j)\big).
\end{equation}
\begin{equation}
\mathcal{I}_3(i,j)=\mathrm{z}\!\big(R_{d^{(2)}}(i,j)\big).
\end{equation}

\section{DINOv3 tuning pipeline}
Here, we provide the equations and description for how we use the DINOv3 model for regression to predict a scalar value.
In the previous section, we converted our 1D signals to 3-channel 2D signals, making them suitable for feeding to RGB channel-expecting VFMs.

\paragraph{1) 1D to 2D conversion.}
As seen previously, we first arrive at the triplet: $(\mathcal{I}_1,\mathcal{I}_2,\mathcal{I}_3)$ using a chosen input representation scheme.

\paragraph{2) Match ImageNet stats.}
After the regular z-score normalization in the previous step, we apply the ImageNet mean/std (per channel).
\begin{equation}
\hat I_c(f,t)=\frac{I(f,t)-\mu^{\mathrm{IM}}_c}{\sigma^{\mathrm{IM}}_c}\ \ \ (c=1,2,3).
\end{equation}
Here $\mu_L,\sigma_L$ are the mean/std over $(f,t)$; $\mu^{\mathrm{IM}}=(0.485,0.456,0.406)$ and\\
$\sigma^{\mathrm{IM}}=(0.229,0.224,0.225)$.

\paragraph{3) ViT patching for our ViT-B/16.}
For $F$x$T$ features, we resize and pad them to multiples of the patch size $p{=}16$ to get the $N$ patches.
\begin{equation}
H_t=\Big\lceil \tfrac{F}{16}\Big\rceil 16,\qquad
W_t=\Big\lceil \tfrac{T}{16}\Big\rceil 16,\qquad
H_f=\tfrac{H_t}{16},\quad W_f=\tfrac{W_t}{16},\quad N=H_fW_f.
\end{equation}

\paragraph{4) Patch embedding to the transformer width.}
Each $3{\times}16{\times}16$ patch (flattened) is projected to $D{=}768$, which is the ViT-B inner dimension.
For $W_E$ being the projector and $u_i$ the patch (combined from all 3 channels),
\begin{equation}
\mathbf{x}_i=\mathbf{W}_E\,\mathbf{u}_i+\mathbf{b}_E,\qquad
\mathbf{u}_i\in\mathbb{R}^{3\cdot 16^2},\ \mathbf{x}_i\in\mathbb{R}^{D}.
\end{equation}

\paragraph{5) Token sequence with CLS and register tokens.}
Now, we demonstrate the transformer processing of all congregated tokens.
DINOv3 adds $R$ learned \emph{register tokens} (in addition to a CLS), then adds positions:
\begin{equation}
\mathbf{Z}^{(0)}=\big[\mathbf{x}_{\mathrm{cls}},\,\mathbf{r}_1,\ldots,\mathbf{r}_R,\,\mathbf{x}_1,\ldots,\mathbf{x}_N\big]+\mathbf{P}
\ \in\ \mathbb{R}^{(1+R+N)\times D}.
\end{equation}

\paragraph{6) Recovering essential output tokens.}
After the transformer, we discard CLS and registers, keep patch tokens, and reshape to a feature map.
\begin{equation}
\mathbf{F}\in\mathbb{R}^{D\times H_f\times W_f}.
\end{equation}

\paragraph{7) Mask-aware soft attention pooling.}
We now have a simple attentive pooling.
We compute a score per location, apply softmax over valid positions (masked as $M$, where data is valid and zeros are not padded), and then take a weighted sum.
\begin{equation}
s(h,w)=\mathbf{w}_s^{\top}\mathbf{F}_{:,h,w},\qquad
Z=\sum_{h',w'} e^{s(h',w')} M(h',w').
\end{equation}
\begin{equation}
\alpha(h,w)=\frac{e^{s(h,w)}\,M(h,w)}{Z},\qquad
\mathbf{p}=\sum_{h,w}\alpha(h,w)\,\mathbf{F}_{:,h,w}\in\mathbb{R}^{D}.
\end{equation}

\paragraph{8) Final regression head.}
We use a simple regression head with LN (LayerNorm), GELU activation, and standard weights and biases terms to map to the final estimand.
\begin{equation}
\hat y=\mathbf{w}_2^{\top}\!\Big(\mathrm{GELU}\big(\mathbf{W}_1\,\mathrm{LN}(\mathbf{p})+\mathbf{b}_1\big)\Big)+b_2.
\end{equation}

In summary, the features are ``patchified'', normalized, and fed with CLS and register tokens. Later, main features are pooled and projected to predict vital signs (BP, Heart rate (HR), Respiration Rate (RR), and SPO2/oxygen saturation) and laboratory tests.

\section{Differences in SIGLIP-2 tuning pipeline}
In this section, we will note differences in pre-processing for SIGLIP-2 VFM.
The majority of the processing is the same as the previous one.

\paragraph{1) Normalization .}
Here, we apply CLIP~\cite{radford2021learning}/SigLIP-style channel normalization
(mean $=0.5$, std $=0.5$) \emph{after} the same per-sample z-normalization.
\begin{equation}
\hat I^{\text{sig}}_c(f,t) = \frac{I_c(f,t) - 0.5}{0.5}, \qquad c \in \{1,2,3\}.
\end{equation}

\paragraph{2) Patching.}
SIGLIP-2 (vision tower) uses patch size $p=14$ (vs.\ 16 in ViT-B/16 for DINOv3), so we do the following to produce $N$ patches.
\begin{equation}
p=14, \qquad
H_t=\left\lceil \frac{F}{14} \right\rceil 14, \quad
W_t=\left\lceil \frac{T}{14} \right\rceil 14.
\end{equation}
\begin{equation}
H_f=\frac{H_t}{14}, \qquad W_f=\frac{W_t}{14}, \qquad N = H_f W_f.
\end{equation}

\paragraph{3) Input token sequence.}
SIGLIP-2 vision typically has \emph{no register tokens}, so we do the following. Later, after transformer processing, we drop the CLS token as per the previous protocol.
\begin{equation}
\mathbf{Z}^{(0)} = [\,\mathbf{x}_{\text{cls}},\,\mathbf{x}_{1},\ldots,\mathbf{x}_{N}\,] + \mathbf{P}
\ \in\ \mathbb{R}^{(1+N) \times D}.
\end{equation}

\section{Results and discussion}
\label{sec:RESULTS}

We present results in three tables (Tables~\ref{tab:main}-\ref{tab:alternate}).
In Table~\ref{tab:main}, we detail the BP prediction results on the seven blood pressure datasets.
The results are in the Mean Absolute Error (MAE) metric in mmHg standard pressure units.
VFMs are compared with the generalist time-series FM MOMENT 385M and PPG FM PPG-GPT 345M.
The large baseline FMs are fully tuned to achieve optimal performance, making it challenging for VFMs to compete with a limited tuning budget.
VFMs are significantly smaller than baseline FMs and are partially tuned using LoRA on attention layers, with additional last-layer tuning.
VFMs wins 9/14 tasks (DINOv3: 3/14, SIGLIP2: 6/14), while generalist time-series FM MOMENT wins 1/14 and PPG FM wins 4/14.
This highlights the power of VFMs even when compared to PPG FM.

\begin{table}[t]
\centering
\small
\caption{Main results (MAE; lower is better). Slash entries report DBP/SBP error}
\begin{tabular}{l|c|c|c|c}
\toprule
\textbf{BP dataset} & \textbf{MOMENT} & \textbf{PPG-GPT} & \textbf{DINOv3} & \textbf{SigLIP-2} \\
\midrule
PPG-BP     & 9.17/17.07 & 10.27/\textbf{16.25} & 9.14/16.42 & \textbf{8.71}/16.72 \\
\hline
Aurora-O   & 21.11/25.11 & 7.96/20.44 & 5.71/18.98 & \textbf{5.32}/\textbf{18.57} \\
\hline
Aurora-A   & 13.79/13.82 & 9.74/13.77 & 9.51/13.58 & \textbf{9.45}/\textbf{13.35} \\
\hline
CAS-BP     & 8.39/14.39  & 8.49/14.34 & \textbf{8.31}/\textbf{13.04} & 8.54/13.22 \\
\hline
Vital Vids & 7.95/18.19  & \textbf{7.93}/18.30 & 9.10/18.07 & 8.21/\textbf{17.77} \\
\hline
BCG        & \textbf{5.86}/4.46   & 7.32/\textbf{4.31}  & 9.54/5.12  & 7.98/6.51  \\
\hline
BUT-PPG    & 8.46/33.00  & \textbf{7.86}/15.53 & 7.92/\textbf{15.07} & 8.04/15.37 \\
\bottomrule
\end{tabular}
\label{tab:main}
\end{table}

\begin{table}[t]
\centering
\small
\caption{Task-wise errors. Units per row: Heart rate (BPM), Respiratory rate (BRPM), SpO$_2$ (\%), Sodium (MEQ/L), Potassium (MEQ/L), Lactate (MMOL/L).}
\begin{tabular}{l|c|c|c|c}
\toprule
\textbf{Estimand} & \textbf{MOMENT} & \textbf{PPG-GPT} & \textbf{DINOv3} & \textbf{SigLIP-2} \\
\midrule
Heart rate  & 10.59 & \textbf{7.83} & 8.10 & 8.27 \\
\hline
Resp. rate  & 4.63  & 4.47 & \textbf{4.41} & 4.42 \\
\hline
SPO$_2$     & 2.24  & 2.28 & \textbf{2.19} & 2.20 \\
\hline
Sodium      & \textbf{3.63}  & 3.74 & 3.73 & 3.74 \\
\hline
Potassium   & 0.47  & \textbf{0.46} & \textbf{0.46} & \textbf{0.46} \\
\hline
Lactate     & \textbf{1.15}  & 1.19 & 1.22 & 1.24 \\
\bottomrule
\end{tabular}
\label{tab:lab}
\end{table}

In Table~\ref{tab:lab}, we note the performance on additional tasks.
For vital tasks, we have heart rate, respiration rate, and SPO2 estimation.
For a more fine-grained waveform analysis task, \textit{blood lab value estimation}, we choose sodium, potassium, and lactate.
This time, VFMs are tied to baseline FMs, with each winning 4/8 tasks.
However, individually, a VFM (DINOv3) remains the best (3/8).
By reporting on heterogeneous and diverse tasks, we demonstrate the generality of our approach.
Note that all tasks here are out-of-domain, i.e., the test and pre-training data do not match for all models.
This indicates that the baseline FMs generalize to unknown domains, although they potentially lack the stronger 3D representations observed in VFMs.

In Table~\ref{tab:alternate}, we show that the performance of our approach can be further improved by exploring more 3D representations.
On a subset of selected tasks, VFMs win 5/6 tasks - vastly outperforming baseline FMs (1/6).
STFT, STFT+phase, and recurrence features achieve success rates of 2/6, 2/6, and 1/6, respectively, across the tasks, with notable SBP MAE reduction by recurrence features for the PPG-BP dataset.
This suggests that combining different features may enable consistent outperformance.
We again defer such investigation of feature pooling and/or learned fusion to future work.

We want to highlight that prior works have explored more 3D features to bring additional improvements: wavelet scalogram, Gramian Angular Fields, Markov Transition Fields, and (multi-scale) recurrence plots, as highlighted in Section~\ref{sec:introduction}.
Through our work, we provide a proof-of-concept for the general PPG analysis capability of VFMs.
In the future, we can expand our study to investigate the emergent capability of FMs from other domains.
Effective PPG processing has vast implications for wearable consumer health as well as triage/risk assessment by a medical facility.
We believe stronger feature/modality fusion systems~\cite{sun2024coronary} can be obtained through exploration of additional PPG analysis tools.

\begin{table}[!h]
\centering
\small
\caption{Alternate PPG representations with DINOv3. Slash entries report DBP/SBP error}
\begin{tabular}{l|c|c|c|c|c}
\toprule
\textbf{BP dataset} & \textbf{MOMENT} & \textbf{PPG-GPT} & \textbf{STFT} & \textbf{STFT+phase} & \textbf{Recurrence} \\
\midrule
PPG-BP     & 9.17/17.07 & 10.27/16.25 & 9.14/16.42 & \textbf{8.71}/16.75 & 9.04/\textbf{13.22} \\
\hline
Aurora-O   & 21.11/25.11 & 7.96/20.44 & \textbf{5.71}/\textbf{18.98} & 5.80/19.28 & 6.42/19.64 \\
\hline
Vital Vids & 7.95/18.19 & \textbf{7.93}/18.30  & 9.10/18.07 & 8.05/\textbf{17.43} & 8.04/18.06 \\
\bottomrule
\end{tabular}
\label{tab:alternate}
\end{table}

\section{Conclusion}
\label{sec:CONCLUSIONS}
In our work Vision4FM, we comprehensively evaluate the emergent capability of Vision Foundation Models trained on vastly different data to accept 3D ``imagified'' representations of PPG signals for various downstream tasks.
In contrast to prior work, we focus on comprehensive evaluation, selection of strong baseline FMs for comparison, and unimodal PPG input.
We also provide a simple recipe for using state-of-the-art VFMs to enable further research.
We primarily investigated cuff-less BP estimation, and also reported results on other vital signs estimation and blood lab estimation tasks.
VFMs resulted in the best performance on the majority of the tasks when compared to baseline time-series FM and PPG FM.
This has significant clinical implications, as this may pave the way for feature fusion methods from various FMs available at the bedside.
Through our exploration of phase and recurrence features, we demonstrate the potential complementarity of such features due to observed scattered improvements.
In the future, we will report the scalability of Vision4PPG with bigger VFMs.
We will also explore the generalizability of domain-specific vision models, such as those for satellite images, which are often released separately.

\section*{Availability of software code}
\label{sec:AVAILABILITY}
Our software code is available at the following URL:\\
\url{https://github.com/saurabh-kataria/Vision4PPG}

\footnotesize
\bibliographystyle{unsrt}
\bibliography{bibliography_CIBB_file.bib} 
\normalsize

\end{document}